\documentclass[10pt,twocolumn,letterpaper]{article}

\usepackage{wacv}
\usepackage{times}
\usepackage{epsfig}
\usepackage{graphicx}
\usepackage{amsmath}
\usepackage{amssymb}
\usepackage{bm} 
\makeatletter
\def\hlinew#1{%
	\noalign{\ifnum0=`}\fi\hrule \@height #1 \futurelet
	\reserved@a\@xhline}
\makeatother

\wacvfinalcopy 

\def\wacvPaperID{****} 

\ifwacvfinal\fi
\begin{document}

\title{DeepLung: Deep 3D Dual Path Nets for \\ Automated Pulmonary Nodule Detection and Classification}

\author{Wentao Zhu$^{1}$ \hspace{2cm} Chaochun Liu$^{2}$ \hspace{2cm} Wei Fan$^{3}$ \hspace{2cm} Xiaohui Xie$^{1}$ \\
$^1$University of California, Irvine \hspace{2cm} $^2$Baidu Research \hspace{2cm} $^3$Tencent Medical AI Lab \\
{\tt\small \{wentaoz1, xhx\}@ics.uci.edu} \hspace{2cm} {\tt\small liuchaochun@baidu.com} \hspace{2cm} {\tt\small davidwfan@tencent.com}}

\maketitle
\ifwacvfinal\thispagestyle{empty}\fi

\begin{abstract}
In this work, we present a fully automated lung computed tomography (CT) cancer diagnosis system, DeepLung. DeepLung consists of two components, nodule detection (identifying the locations of candidate nodules) and classification (classifying candidate nodules into benign or malignant). Considering the 3D nature of lung CT data and the compactness of dual path networks (DPN), two deep 3D DPN are designed for nodule detection and classification respectively. Specifically, a 3D Faster Regions with Convolutional Neural Net (R-CNN) is designed for nodule detection with 3D dual path blocks and a U-net-like encoder-decoder structure to effectively learn nodule features. For nodule classification, gradient boosting machine (GBM) with 3D dual path network features is proposed. The nodule classification subnetwork was validated on a public dataset from LIDC-IDRI, on which it achieved better performance than state-of-the-art approaches and surpassed the performance of experienced doctors based on image modality. Within the DeepLung system, candidate nodules are detected first by the nodule detection subnetwork, and nodule diagnosis is conducted by the classification subnetwork. Extensive experimental results demonstrate that DeepLung has performance comparable to experienced doctors both for the nodule-level and patient-level diagnosis on the LIDC-IDRI dataset.\footnote{https://github.com/uci-cbcl/DeepLung.git}
\end{abstract}

\begin{figure*}
	\begin{center}
		\includegraphics[width=\linewidth]{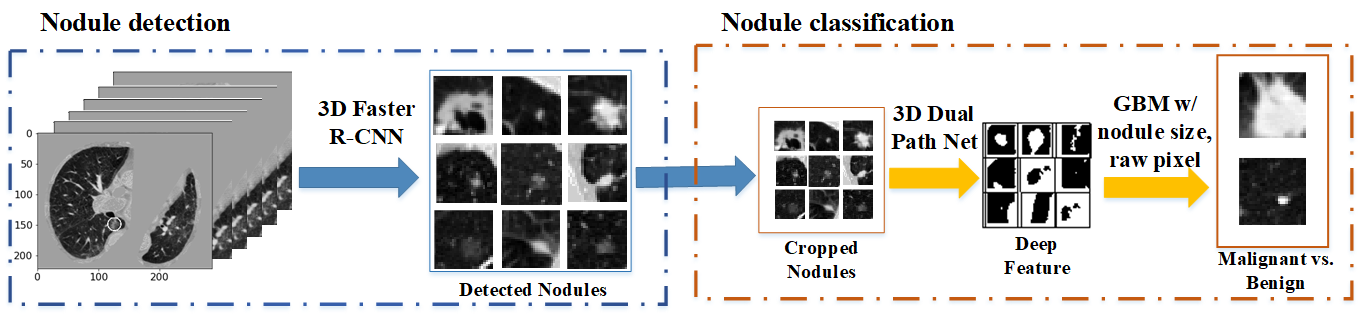}
		\caption{The framework of DeepLung. DeepLung first employs 3D Faster R-CNN to generate candidate nodules. Then it uses deep 3D DPN to extract deep features from the detected and cropped nodules. Lastly, GBM with deep features, detected nodule size, and raw pixels is employed for classification. Patient-level diagnosis can be achieved by fusing the classification results of detected nodules in the CT.}\label{fig:framework}
	\end{center}
\end{figure*}
\section{Introduction}

Lung cancer is the most common cause of cancer-related death in men. Low-dose lung CT screening provides an effective way for early diagnosis, which can sharply reduce the lung cancer mortality rate. Advanced computer-aided diagnosis systems (CADs) are expected to have high sensitivities while at the same time maintaining low false positive rates. Recent advances in deep learning enable us to rethink the ways of clinician lung cancer diagnosis.

Current lung CT analysis research mainly includes nodule detection \cite{dou2017automated,ding2017accurate}, and nodule classification \cite{shen2015multi,dlcls,hussein2017risk,yan2016classification}. There is few work on building a complete lung CT cancer diagnosis system for fully automated lung CT cancer diagnosis using deep learning, integrating both nodule detection and nodule classification. It is worth exploring a whole lung CT cancer diagnosis system and understanding how far the performance of current deep learning technology differs from that of experienced doctors. To our best knowledge, this is the first work for a fully automated and complete lung CT cancer diagnosis system using deep nets. 

The emergence of large-scale dataset, LUNA16 \cite{setio2016pulmonary}, accelerated the nodule detection related research. Typically, nodule detection consists of two stages, region proposal generation and false positive reduction. Traditional approaches generally require manually designed features such as morphological features, voxel clustering and pixel thresholding \cite{murphy2009large,jacobs2014automatic}. Recently, deep ConvNets, such as Faster R-CNN \cite{kaimingfasterrcnn,liao2017evaluate} and fully ConvNets \cite{long2015fully,zhu2017adversarial,zhe_vqa,zhe_scene,zhe_action}, are employed to generate candidate bounding boxes \cite{ding2017accurate,dou2017automated}. In the second stage, more advanced methods or complex features, such as carefully designed texture features, are used to remove false positive nodules. Because of the 3D nature of CT data and the effectiveness of Faster R-CNN for object detection in 2D natural images \cite{iccv17detectorcompare}, we design a 3D Faster R-CNN for nodule detection with 3D convolutional kernels and a U-net-like encoder-decoder structure to effectively learn latent features \cite{ronneberger2015u}. The U-Net structure is basically a convolutional autoencoder, augmented with skip connections between encoder and decoder layers \cite{ronneberger2015u}. Although it has been widely used in the context of semantic segmentation, being able to capture both contextual and local information should be very helpful for nodule detections as well. Because 3D ConvNet has too large a number of parameters and is difficult to train on public lung CT datasets of relatively small sizes, 3D dual path network is employed as the building block since deep dual path network is more compact and provides better performance than deep residual network at the same time \cite{chen2017dual}. 

Before the era of deep learning, manual feature engineering followed by classifiers was the general pipeline for nodule classification \cite{han2013texture}. After the large-scale LIDC-IDRI \cite{armato2011lung} dataset became publicly available, deep learning-based methods have become the dominant framework for nodule classification research \cite{dlcls,zhu2016co}. Multi-scale deep ConvNet with shared weights on different scales has been proposed for the nodule classification \cite{shen2015multi}. The weight sharing scheme reduces the number of parameters and forces the multi-scale deep ConvNet to learn scale-invariant features. Inspired by the recent success of dual path network (DPN) on ImageNet \cite{chen2017dual,imagenet}, we propose a novel framework for CT nodule classification. First, we design a deep 3D dual path network to extract features. As gradient boosting machines (GBM) are known to have superb performance given effective features, we use GBM with deep 3D dual path features, nodule size, and cropped raw nodule CT pixels for the nodule classification \cite{gbt}. 


Finally, we built a fully automated lung CT cancer diagnosis system, henceforth called DeepLung, by combining the nodule detection network and nodule classification network together, as illustrated in Fig. \ref{fig:framework}. For a CT image, we first use the detection subnetwork to detect candidate nodules. Next, we employ the classification subnetwork to classify the detected nodules into either malignant or benign. Finally, the patient-level diagnosis result can be achieved for the whole CT by fusing the diagnosis result of each nodule.

Our main contributions are as follows: 1) To fully exploit the 3D CT images, two deep 3D ConvNets are designed for nodule detection and classification respectively. Because 3D ConvNet contains too many parameters and is difficult to train on relatively small public lung CT datasets, we employ 3D dual path networks as the neural network architecture since DPN uses less parameters and obtains better performance than residual network \cite{chen2017dual}. Specifically, inspired by the effectiveness of Faster R-CNN for object detection \cite{iccv17detectorcompare}, we propose 3D Faster R-CNN for nodule detection based on 3D dual path network and U-net-like encoder-decoder structure, and deep 3D dual path network for nodule classification. 2) Our classification framework achieves better performance compared with state-of-the-art approaches, and surpasses the performance of experienced doctors on the public dataset, LIDC-IDRI. 3) Our fully automated DeepLung system, nodule classification based on detection, is comparable to the performance of experienced doctors both on nodule-level and patient-level diagnosis. 

\section{Related Work}

Traditional nodule detection involves hand-designed features or descriptors \cite{lopez2015large} requiring domain expertise. Recently, several works have been proposed to use deep ConvNets for nodule detection to automatically learn features, which is proven to be much more effective than hand-designed features. Setio et al. proposes multi-view ConvNet for false positive nodule reduction \cite{setio2016pulmonarymultiview}. Due to the 3D nature of CT scans, some work proposed 3D ConvNets to handle the challenge. The 3D fully ConvNet (FCN) is proposed to generate region candidates, and deep ConvNet with weighted sampling is used for false positive reduction \cite{dou2017automated}. Ding et al. and Liao et al. use the Faster R-CNN to generate candidate nodules followed by 3D ConvNets to remove false positive nodules \cite{ding2017accurate,liao2017evaluate}. Due to the effective performance of Faster R-CNN \cite{iccv17detectorcompare,kaimingfasterrcnn}, we design a novel network, 3D Faster R-CNN with 3D dual path blocks, for the nodule detection. Further, a U-net-like encoder-decoder scheme is employed for 3D Faster R-CNN to effectively learn the features \cite{ronneberger2015u}.

Nodule classification has traditionally been based on segmentation \cite{el20113d} and manual feature design \cite{aerts2014decoding}. Several works designed 3D contour feature, shape feature and texture feature for CT nodule diagnosis \cite{way2006computer,el20113d,han2013texture}. Recently, deep networks have been shown to be effective for medical images. Artificial neural network was implemented for CT nodule diagnosis \cite{suzuki2005computer}. More computationally effective network, multi-scale ConvNet with shared weights for different scales to learn scale-invariant features, is proposed for nodule classification \cite{shen2015multi}. Deep transfer learning and multi-instance learning is used for patient-level lung CT diagnosis \cite{dlcls,zhu2017deep}. A comparative study on 2D and 3D ConvNets is conducted and 3D ConvNet is shown to be better than 2D ConvNet for 3D CT data \cite{yan2016classification}. Furthermore, a multi-task learning and transfer learning framework is proposed for nodule diagnosis \cite{hussein2017risk}. Different from their approaches, we propose a novel classification framework for CT nodule diagnosis. Inspired by the recent success of deep dual path network (DPN) on ImageNet \cite{chen2017dual}, we design a novel 3D DPN to extract features from raw CT nodules. In part to the superior performance of GBM with complete features, we employ GBM with different levels of granularity ranging from raw pixels, DPN features, to global features such as nodule size for the nodule diagnosis. Patient-level diagnosis can be achieved by fusing the nodule-level diagnosis. 

\section{DeepLung Framework}
Our fully automated lung CT cancer diagnosis system consists of two parts: nodule detection and classification. We design a 3D Faster R-CNN for nodule detection, and propose GBM with deep 3D DPN features, raw nodule CT pixels and nodule size for nodule classification.

\subsection{3D Faster R-CNN with Deep 3D Dual Path Net for Nodule Detection}
\begin{figure}
	\includegraphics[width=\linewidth]{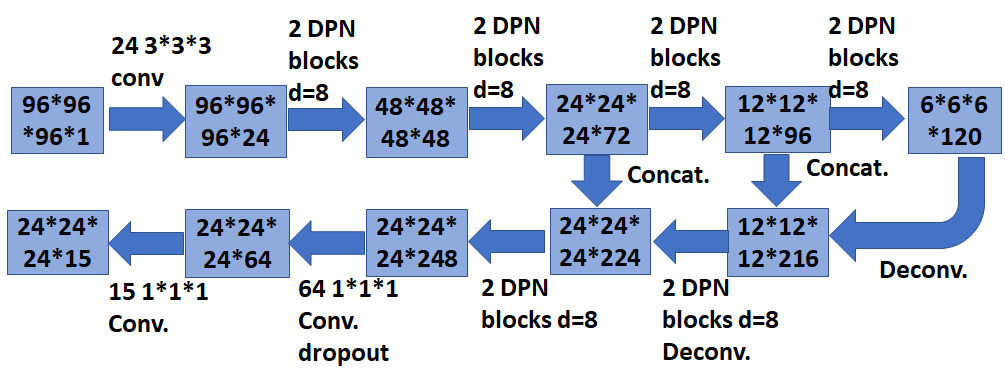}
	\caption{The 3D Faster R-CNN framework contains 3D dual path blocks and a U-net-like encoder-decoder structure. We design 26 layers 3D dual path network for the encoder subnetwork. The model employs 3 anchors and multi-task learning loss, including coordinates \((x, y, z)\) and diameter \(d\) regression, and candidate box classification. The numbers in boxes are feature map sizes in the format (\(\#\)slices*\(\#\)rows*\(\#\)cols*\(\#\)maps). The numbers above the connections are in the format (\(\#\)filters \quad \(\#\)slices*\(\#\)rows*\(\#\)cols).}
	\label{fig:3dfasterrcnn}
\end{figure}

Inspired by the success of dual path network on the ImageNet \cite{chen2017dual,imagenet}, we design a deep 3D DPN framework for lung CT nodule detection and classification in Fig. \ref{fig:3dfasterrcnn} and Fig. \ref{fig:dpncls}. 
\begin{figure}
	\begin{center}
		\includegraphics[width=\linewidth]{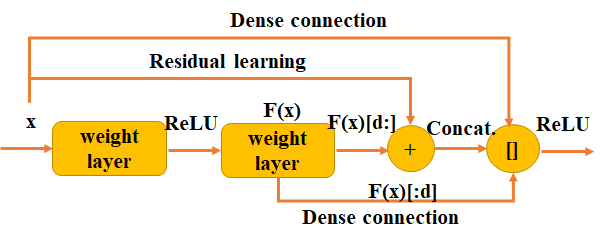}
		\caption{Illustration of dual path connection \cite{chen2017dual}, which benefits both from the advantage of residual learning \cite{he2016deep} and that of dense connection \cite{huang2016densely} from network structure design intrinsically.}
		\label{fig:dualpath}
	\end{center}
\end{figure}
Dual path connection benefits both from the advantage of residual learning and that of dense connection \cite{he2016deep,huang2016densely}. The shortcut connection in residual learning is an effective way to eliminate vanishing gradient phenomenon in very deep networks. From a learned feature sharing perspective, residual learning enables feature reuse, while dense connection has an advantage of exploiting new features \cite{chen2017dual}. Additionally, densely connected network has fewer parameters than residual learning because there is no need to relearn redundant feature maps. The assumption of dual path connection is that there might exist some redundancy in the exploited features. And dual path connection uses part of feature maps for dense connection and part of them for residual learning. In implementation, the dual path connection splits its feature maps into two parts. One part, \({\bf{F}(\bf{x})}[d:]\), is used for residual learning, the other part, \({\bf{F}(\bf{x})}[:d]\), is used for dense connection as shown in Fig. \ref{fig:dualpath}. Here \(d\) is a hyper-parameter for deciding how many new features to be exploited. The dual path connection can be formulated as
\begin{equation}
{\bf{y}} = {\bf{G}}([{\bf{x}}[:d], {\bf{F}(\bf{x})}[:d], {\bf{F}(\bf{x})}[d:]+\bf{x}[d:]),
\end{equation}
where \(\bf{y}\) is the feature map for dual path connection, \(\bf{G}\) is used as ReLU activation function, \(\bf{F}\) is convolutional layer functions, and \(\bf{x}\) is the input of dual path connection block. Dual path connection integrates the advantages of the two advanced frameworks, residual learning for feature reuse and dense connection for the ability to exploit new features, into a unified structure which obtained success on the ImageNet dataset\cite{imagenet}. We design deep 3D neural nets based on 3D DPN because of its compactness and effectiveness.

The 3D Faster R-CNN with a U-net-like encoder-decoder structure and 3D dual path blocks is illustrated in Fig. \ref{fig:3dfasterrcnn}. Due to the GPU memory limitation, the input of 3D Faster R-CNN is cropped from 3D reconstructed CT images with pixel size \(96 \times 96 \times 96\). The encoder network is derived from 2D DPN \cite{chen2017dual}. Before the first max-pooling, two convolutional layers are used to generate features. After that, eight dual path blocks are employed in the encoder subnetwork. We integrate the U-net-like encoder-decoder design concept in the detection to learn the deep nets efficiently \cite{ronneberger2015u}. In fact, for the region proposal generation, the 3D Faster R-CNN conducts pixel-wise multi-scale learning and the U-net is validated as an effective way for pixel-wise labeling. This integration makes candidate nodule generation more effective. In the decoder network, the feature maps are processed by deconvolution layers and dual path blocks, and are subsequently concatenated with the corresponding layers in the encoder network \cite{zeiler2010deconvolutional}. Then a convolutional layer with dropout (dropout probability 0.5) is used in the second to the last layer. In the last layer, we design 3 anchors, 5, 10, 20, for scale references which are designed based on the distribution of nodule sizes. For each anchor, there are 5 parts in the loss function, classification loss \(L_{cls}\) for whether the current box is a nodule or not, regression loss \(L_{reg}\) for nodule coordinates \(x, y, z\) and nodule size \(d\).

If an anchor overlaps a ground truth bounding box with the intersection over union (IoU) higher than 0.5, we consider it as a positive anchor (\(p^{\star}=1\)). On the other hand, if an anchor has IoU with all ground truth boxes less than 0.02, we consider it as a negative anchor (\(p^{\star}=0\)). The multi-task loss function for the anchor \(i\) is defined as
\begin{equation}
L(p_i, {\bf{t}_i}) = \lambda L_{cls}(p_i, p_i^{\star}) + p_i^{\star} L_{reg}({\bf{t}_i}, {\bf{t}_i}^{\star}),
\end{equation}
where \(p_i\) is the predicted probability for current anchor \(i\) being a nodule, \(\bf{t}_i\) is the predicted relative coordinates for nodule position, which is defined as
\begin{equation}
{\bf{t}_i} = (\frac{x-x_a}{d_a}, \frac{y-y_a}{d_a}, \frac{z-z_a}{d_a}, \log(\frac{d}{d_a})),
\end{equation}
where \((x, y, z, d)\) are the predicted nodule coordinates and diameter in the original space, \((x_a, y_a, z_a, d_a)\) are the coordinates and scale for the anchor \(i\). For ground truth nodule position, it is defined as
\begin{equation}
{{\bf{t}}_i^{\star}} = (\frac{x^{\star}-x_a}{d_a}, \frac{y^{\star}-y_a}{d_a}, \frac{z^{\star}-z_a}{d_a}, \log(\frac{d^{\star}}{d_a})),
\end{equation}
where \((x^{\star}, y^{\star}, z^{\star}, d^{\star})\) are nodule ground truth coordinates and diameter. The \(\lambda\) is set as \(0.5\). For \(L_{cls}\), we used binary cross entropy loss function. For \(L_{reg}\), we used smooth \(l_1\) regression loss function \cite{girshick2015fast}. 
\subsection{Gradient Boosting Machine with 3D Dual Path Net Feature for Nodule Classification}

\begin{figure}
	\begin{center}
		\includegraphics[width=0.9\linewidth]{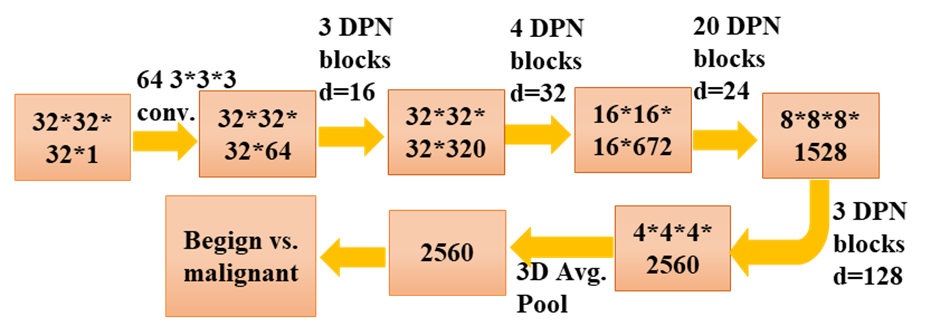}
		\caption{The deep 3D dual path network framework in the nodule classification subnetwork, which contains 30 3D dual path connection blocks. After the training, the deep 3D dual path network feature is extracted for gradient boosting machine to do nodule diagnosis. The numbers are of the same formats as Fig. \ref{fig:3dfasterrcnn}.}
		\label{fig:dpncls}
	\end{center}
\end{figure}

For CT data, advanced method should be effective to extract 3D volume feature \cite{yan2016classification}. We design a 3D deep dual path network for the 3D CT lung nodule classification in Fig. \ref{fig:dpncls}. The main reason we employ dual modules for detection and classification is that classifying nodules into benign and malignant requires the system to learn finer-level features, which can be achieved by focusing only on nodules. In addition, it allows to introduce extra features in the final classification. We first crop CT data centered at predicted nodule locations with size \(32\times 32 \times 32\). After that, a convolutional layer is used to extract features. Then 30 3D dual path blocks are employed to learn higher level features. Lastly, the 3D average pooling and binary logistic regression layer are used for benign or malignant diagnosis. 

The deep 3D dual path network can be used as a classifier for nodule diagnosis directly and it can also be employed to learn effective features. 
We construct feature by concatenating the learned deep 3D DPN features (the second from the last layer (2,560 dimension)), nodule size, and raw 3D cropped nodule pixels. Given complete and effective features, GBM is a superb method to build an advanced classifier \cite{gbt}. We validate the feature combining nodule size with raw 3D cropped nodule pixels in combination with the GBM classifier and obtained \(86.12 \%\) average test accuracy. Lastly, we employ GBM with the constructed feature and achieve the best diagnosis performance. 

\subsection{DeepLung System: Fully Automated Lung CT Cancer Diagnosis}
The DeepLung system includes the nodule detection using the 3D Faster R-CNN and nodule classification using GBM with constructed feature (deep 3D dual path features, nodule size and raw nodule CT pixels) as shown in Fig. \ref{fig:framework}. 

Due to the GPU memory limitation, we first split the whole CT into several \(96 \times 96 \times 96\) patches, process them through the detector, and combine the detected results together. We only keep the detected boxes of detection probabilities larger than 0.12 (threshold as -2 before sigmoid function). After that, non-maximum suppression (NMS) is adopted based on detection probability with the intersection over union (IoU) threshold as 0.1. Here we expect to not miss too many ground truth nodules. 

After we get the detected nodules, we crop the nodule with the center as the detected center and size of \(32 \times 32 \times 32\). The detected nodule size is kept as a feature input for later downstream classification. The deep 3D DPN is employed to extract features. We use the GBM and construct features to conduct diagnosis for the detected nodules. For pixel feature, we use the cropped size of \(16 \times 16 \times 16\) and center as the detected nodule center in the experiments. For patient-level diagnosis, if one of the detected nodules is positive (cancer), the patient is classified as having cancer. Conversely, if all detected nodules are negative, the patient is considered non-cancer.

\section{Experiments}
We conduct extensive experiments to validate the DeepLung system. We perform 10-fold cross validation using the detector on LUNA16 dataset. For nodule classification, we use the LIDC-IDRI annotation, and employ the LUNA16's patient-level dataset split. Finally, we also validate the whole system based on the detected nodules both on patient-level diagnosis and nodule-level diagnosis. 

In the training, for each model, we use 150 epochs in total with stochastic gradient descent optimization and momentum as 0.9. The batch size parameter is limited by GPU memory. We use weight decay as \(1 \times 10^{-4}\). The initial learning rate is 0.01, 0.001 after half the total number of epoch, and 0.0001 after epoch 120.

\subsection{Datasets}

LUNA16 dataset is a subset of the largest publicly available dataset for pulmonary nodules, LIDC-IDRI \cite{armato2011lung,setio2016pulmonary}. LUNA16 dataset only has the detection annotations, while LIDC-IDRI contains almost all the related information for low-dose lung CTs including several doctors' annotations on nodule sizes, locations, diagnosis results, nodule texture, nodule margin and other informations. LUNA16 dataset removes CTs with slice thickness greater than 3mm, slice spacing inconsistent or missing slices from LIDC-IDRI dataset, and explicitly gives the patient-level 10-fold cross validation split of the dataset. LUNA16 dataset contains 888 low-dose lung CTs, and LIDC-IDRI contains 1,018 low-dose lung CTs. Note that LUNA16 dataset removes the annotated nodules of size smaller than 3mm. 

For nodule classification, we extract nodule annotations from LIDC-IDRI dataset, find the mapping of different doctors' nodule annotations with the LUNA16's nodule annotations, and obtained the ground truth of nodule diagnosis by averaging different doctors' diagnosis (discarding 0 score for diagnosis which corresponds to N/A.). If the final average score is equal to 3 (uncertain about malignant or benign), we remove the nodule. For the nodules with score greater than 3, we label them as positive. Otherwise, we label them as negative. Because CT slides were annotated by anonymous doctors, the identities of doctors (referred to as Drs 1-4 as the 1st-4th annotations) are not strictly consistent. As such, we refer them as ``simulated'' doctors. To make our results reproducible, we only keep the CTs within LUNA16 dataset, and use the same cross validation split as LUNA16 for classification. 
\subsection{Preprocessing}
Three automated preprocessing steps are employed for the input CT images. First, we clip the raw data into \([-1200, 600]\). Second, we transform the range linearly into \([0, 1]\). Finally, we use LUNA16's given segmentation ground truth and remove the background. 
\subsection{DeepLung for Nodule Detection}
We train and evaluate the detector on LUNA16 dataset following 10-fold cross validation with given patient-level split. In training, we augment the dataset by randomly flipping the image and use cropping scale betweeb 0.75 to 1.25. The evaluation metric, FROC, is the average recall rate at the average number of false positives at 0.125, 0.25, 0.5, 1, 2, 4, 8 per scan, which is the official evaluation metric for LUNA16 dataset \cite{setio2016pulmonary}. In the test phase, we use detection probability threshold as -2 (before sigmoid function), followed by NMS with IoU threshold as 0.1. 

\begin{figure}
	\includegraphics[width=\linewidth]{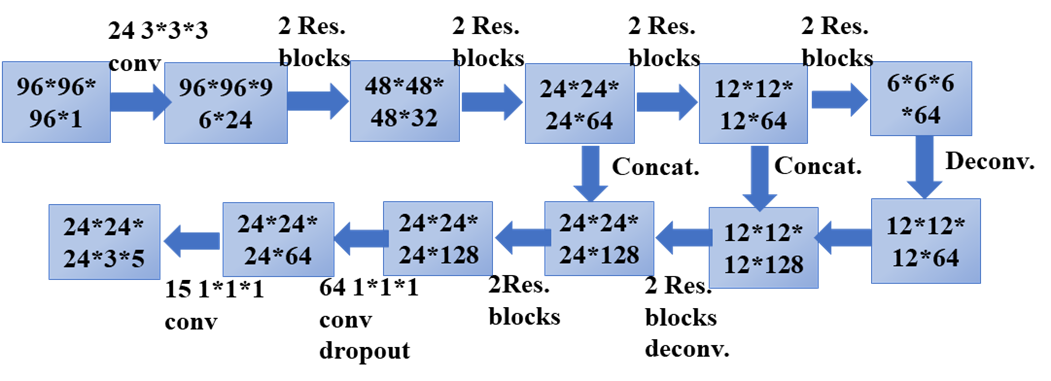}
	\caption{The 3D Faster R-CNN network with 3D residual blocks. It contains several 3D residual blocks. We employ a deep 3D residual network of 18 layers as the encoder subnetwork, which is an extension from 2D Res18 net \cite{he2016deep}.}
	\label{fig:3dfasterrcnncmp}
\end{figure}

To validate the performance of proposed deep 3D dual path network for detection, we employ a deep 3D residual network as a comparison in Fig. \ref{fig:3dfasterrcnncmp}. The encoder part of this baseline network is a deep 3D residual network of 18 layers, which is an extension from 2D Res18 net \cite{he2016deep}. Note that the 3D Res18 Faster R-CNN contains \(5.4\)M trainable parameters, while the 3D DPN26 Faster R-CNN employs \(1.4\)M trainable parameters, which is only \( \frac{\textbf{1}}{\textbf{4}}\) of 3D Res18 Faster R-CNN. 
\begin{figure}
	\begin{center}
		\includegraphics[width=0.9\linewidth]{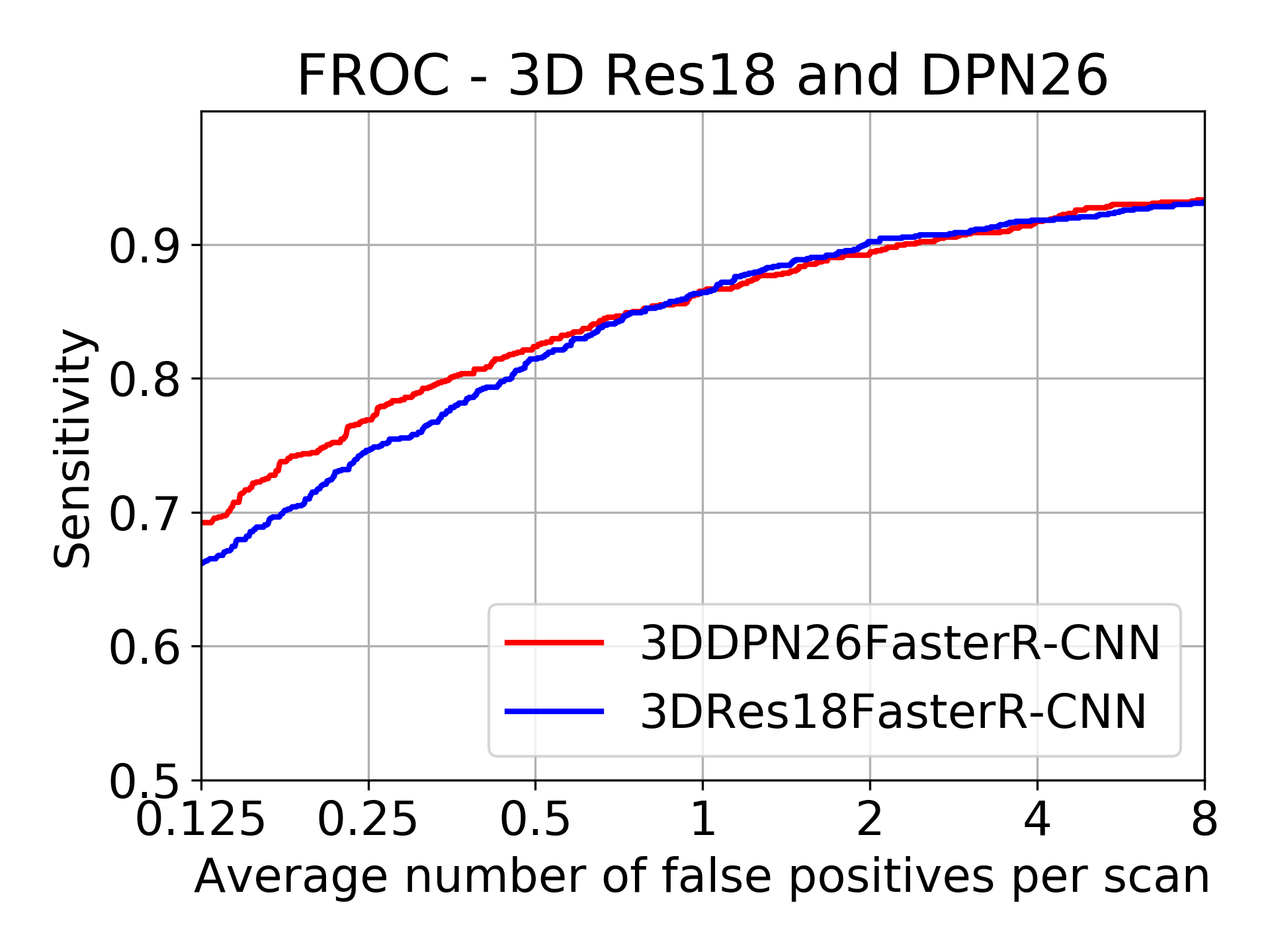}
		\caption{Sensitivity (Recall) rate with respect to false positives per scan. The FROC (average recall rate at the false positives as 0.125, 0.25, 0.5, 1, 2, 4, 8) of 3D Res18 Faster R-CNN is 83.4\%, while the FROC of 3D DPN26 Faster R-CNN is \textbf{84.2\%} with only \( \frac{\textbf{1}}{\textbf{4}}\) of the parameters as 3D Res18 Faster R-CNN. The 3D Res18 Faster R-CNN has a total recall rate 94.6\% for all the detected nodules, while 3D DPN26 Faster R-CNN has a recall rate \textbf{95.8\%}.}
		\label{fig:froc}
	\end{center}
\end{figure}

The FROC performance on LUNA16 is visualized in Fig. \ref{fig:froc}. The solid line is interpolated FROC based on true prediction. The 3D DPN26 Faster R-CNN achieves a FROC score of {\textbf{84.2\%}} without any false positive nodule reduction stage, which is better than the previous 83.9\% using two-stage training \cite{dou2017automated}. The 3D DPN26 Faster R-CNN using only \( \frac{\textbf{1}}{\textbf{4}}\) of the parameters performs better than the 3D Res18 Faster R-CNN, which demonstrates the superior suitability of the 3D DPN for detection. Ding et al. obtains 89.1\% FROC using 2D Faster R-CNN followed by extra false positive reduction classifier \cite{ding2017accurate}, while we only employ enhanced Faster R-CNN with deep 3D dual path for detection. We have recently applied the 3D model to Alibaba Tianchi Medical AI on nodule detection challenge and were able to achieve top accuracy on a hold-out dataset. 

\subsection{DeepLung for Nodule Classification}

We validate the nodule classification performance of the DeepLung system on the LIDC-IDRI dataset with the LUNA16's split principle, 10-fold patient-level cross validation. There are 1,004 nodules of which 450 are positive. In the training, we first pad the nodules of size \(32 \times 32 \times 32\) into \(36 \times 36 \times 36\), randomly crop \(32 \times 32 \times 32\) from the padded data, horizontal flip, vertical flip, z-axis flip the data for augmentation, randomly set \(4\times4\times4\) patch to zero, and normalize the data with the mean and standard deviation obtained from training data. The total number of epochs is 1,050. The initial learning rate is 0.01, and reduce to 0.001 after epoch 525, and finally to 0.0001 after epoch 840. Due to time and resource limitation for training, we use the fold 1, 2, 3, 4, 5 for test, and the final performance is the average performance on the five test folds. The nodule classification performance is concluded in Table \ref{tab:nodcls}.
\begin{table}[]
	\centering
	\caption{Nodule classification comparisons on LIDC-IDRI dataset.}
	\label{tab:nodcls}
	\begin{tabular}{c|c|c}
		\hlinew{0.9pt}
		Models & Accuracy (\%) & Year \\  
		\hlinew{0.7pt}
		Multi-scale CNN \cite{shen2015multi} & 86.84 & 2015 \\ \hline
		Slice-level 2D CNN \cite{yan2016classification} & 86.70 & 2016 \\ \hline
		Nodule-level 2D CNN \cite{yan2016classification} & 87.30 & 2016 \\ \hline
		Vanilla 3D CNN \cite{yan2016classification} & 87.40 & 2016 \\ \hline
		Multi-crop CNN \cite{shen2017multi} & 87.14 & 2017 \\ 
		\hlinew{0.9pt}
		Deep 3D DPN & 88.74 & 2017 \\ \hline
		Nodule Size+Pixel+GBM & 86.12 & 2017 \\ \hline
		All feat.+GBM & \textbf{90.44} & 2017 \\ 
		\hlinew{0.9pt}
	\end{tabular}
\end{table}

From the table \ref{tab:nodcls}, our deep 3D DPN achieves better performance than those of Multi-scale CNN \cite{shen2015multi}, Vanilla 3D CNN \cite{yan2016classification} and Multi-crop CNN \cite{shen2017multi}, because of the strong power of 3D structure and deep dual path network. GBM with nodule size and raw nodule pixels with crop size as \(16 \times 16 \times 16\) achieves comparable performance as multi-scale CNN \cite{shen2015multi} because of the superior classification performance of GBM. Finally, we construct feature with deep 3D dual path network features, 3D Faster R-CNN detected nodule size and raw nodule pixels, and obtain {\bf{90.44\%}} accuracy, which shows the effectiveness of deep 3D dual path network features. 
\subsubsection{Compared with Experienced Doctors on Their Individual Confident Nodules}

We compare our predictions with those of four ``simulated'' experienced doctors on their individually confident nodules (with individual score not 3). Note that about 1/3 annotations are 3. Comparison results are concluded in Table \ref{tab:nodclsdr}.

\begin{table}[]
	\centering
	\caption{Nodule-level diagnosis accuracy (\%) between nodule classification subnetwork in DeepLung and experienced doctors on doctor's individually confident nodules.}
	\label{tab:nodclsdr}
	\begin{tabular}{c|c|c|c|c|c}
		\hline
		& Dr 1 & Dr 2 & Dr 3 & Dr 4 & Average \\ \hline
		Doctors   & 93.44    & 93.69    & 91.82    & 86.03    & 91.25   \\ \hline
		DeepLung & 93.55    & 93.30    & 93.19    & 90.89    & \textbf{92.74}   \\ \hline
	\end{tabular}
\end{table}

From Table \ref{tab:nodclsdr}, these doctors' confident nodules are easy to be diagnosed nodules from the performance comparison between our model's performances in Table \ref{tab:nodcls} and Table \ref{tab:nodclsdr}. To our surprise, the average performance of our model is {\textbf{1.5\%}} better than that of experienced doctors even on their individually confident diagnosed nodules. In fact, our model's performance is better than 3 out of 4 doctors (doctor 1, 3, 4) on the confident nodule diagnosis task. The result validates deep network surpasses human-level performance for image classification \cite{he2016deep}, and the DeepLung is better suited for nodule diagnosis than experienced doctors. 

\begin{table}[]
	\centering
	\caption{Statistical property of predicted malignant probability for borderline nodules (\%)}
	\label{tab:statborderlinenod}
	\begin{tabular}{c|c|c|c|c}
		\hline
		Prediction & \begin{tabular}[c]{@{}l@{}}\(<0.1\) or\\ \(>0.9\)\end{tabular}   &\begin{tabular}[c]{@{}l@{}}\(<0.2\) or\\ \(>0.8\)\end{tabular} & \begin{tabular}[c]{@{}l@{}}\(<0.3\) or\\ \(>0.7\)\end{tabular}  & \begin{tabular}[c]{@{}l@{}}\(<0.4\) or\\ \(>0.6\)\end{tabular}  \\ \hline
		Frequency & 64.98    & 80.14    & 89.75    & 94.80   \\ \hline
	\end{tabular}
\end{table}
We also employ Kappa coefficient, which is a common approach to evaluate the agreement between two raters, to test the agreement between DeepLung and the ground truth \cite{smeeton1985early}. The kappa coefficient of DeepLung is 85.07\%, which is significantly better than the average kappa coefficient of doctors (81.58\%). To evaluate the performance for all nodules including borderline nodules (labeled as 3, uncertain between malignant and benign), we compute the log likelihood (LL) scores of DeepLung and doctors' diagnosis. We randomly sample 100 times from the experienced doctors' annotations as 100 ``simulated'' doctors. The mean LL of doctors is -2.563 with a standard deviation of 0.23. By contrast, the LL of DeepLung is -1.515, showing that the performance of DeepLung is 4.48 standard deviation better than the average performance of doctors, which is highly statistically significant. It is important to analysis the statistical property of predictions for borderline nodules that cannot be conclusively classified by doctors. Interestingly, 64.98\% of the borderline nodules are classified to be either malignant (with probability $>$ 0.9) or benign (with probability $<$ 0.1) in Table \ref{tab:statborderlinenod}.  DeepLung classified most of the borderline nodules of malignant probabilities closer to zero or closer to one, showing its potential as a tool for assisted diagnosis. 

\subsection{DeepLung for Fully Automated Lung CT Cancer Diagnosis}
We also validate the DeepLung for fully automated lung CT cancer diagnosis on the LIDC-IDRI dataset with the same protocol as LUNA16's patient-level split. Firstly, we employ our 3D Faster R-CNN to detect suspicious nodules. Then we retrain the model from nodule classification model on the detected nodules dataset. If the center of detected nodule is within the ground truth positive nodule, it is a positive nodule. Otherwise, it is a negative nodule. Through this mapping from the detected nodule and ground truth nodule, we can evaluate the performance and compare it with the performance of experienced doctors. We adopt the test fold 1, 2, 3, 4, 5 to validate the performance the same as that for nodule classification.

\begin{table}[]
	\centering
	\caption{Comparison between DeepLung's nodule classification on all detected nodules and doctors on all nodules.}
	\label{tab:nodclstpfp}
	\begin{tabular}{c|c|c|c}
		\hline
		Method& TP Set & FP Set & Doctors  \\ \hline
		Acc. (\%)   & 81.42    & 97.02    & 74.05-82.67       \\ \hline
	\end{tabular}
\end{table}

Different from pure nodule classification, the fully automated lung CT nodule diagnosis relies on nodule detection. We evaluate the performance of DeepLung on the detection true positive (TP) set and detection false positive (FP) set individually in Table \ref{tab:nodclstpfp}. If the detected nodule of center within one of ground truth nodule regions, it is in the TP set. If the detected nodule of center out of any ground truth nodule regions, it is in FP set. From Table \ref{tab:nodclstpfp}, the DeepLung system using detected nodule region obtains \(\textbf{81.42\%}\) accuracy for all the detected TP nodules. Note that the experienced doctors obtain 78.36\% accuracy for all the nodule diagnosis on average. The DeepLung system with fully automated lung CT nodule diagnosis still achieves above average performance of experienced doctors. On the FP set, our nodule classification subnetwork in the DeepLung can reduce 97.02\% FP detected nodules, which guarantees that our fully automated system is effective for the lung CT cancer diagnosis. 

\subsubsection{Compared with Experienced Doctors on Their Individually Confident CTs}

We employ the DeepLung for patient-level diagnosis further. If the current CT has one nodule that is classified as positive, the diagnosis of the CT is positive. If all the nodules are classified as negative for the CT, the diagnosis of the CT is negative. We evaluate the DeepLung on the doctors' individually confident CTs for benchmark comparison in Table \ref{tab:ctclsdr}. 
\begin{table}[]
	\centering
	\caption{Patient-level diagnosis accuracy(\%) between DeepLung and experienced doctors on doctor's individually confident CTs.}
	\label{tab:ctclsdr}
	\begin{tabular}{c|c|c|c|c|c}
		\hline
		& Dr 1 & Dr 2 & Dr 3 & Dr 4 & Average \\ \hline
		Doctors   & 83.03    & 85.65    & 82.75    & 77.80    & \textbf{82.31}   \\ \hline
		DeepLung & 81.82    & 80.69    & 78.86    & 84.28    & 81.41   \\ \hline
	\end{tabular}
\end{table}

From Table \ref{tab:ctclsdr}, DeepLung achieves {\bf{81.41\%}} patient-level diagnosis accuracy. This is {\bf{99\%}} of the average performance of four experienced doctors and better than Dr 4 altogether. This performance gives confidence that DeepLung can be a useful tool to assist doctors' in their diagonsis. We further validate our method against the four doctors' individual confidential CTs. The Kappa coefficient of DeepLung is 63.02\%, while the average Kappa coefficient of the doctors is 64.46\%. It implies the predictions of DeepLung are of good agreement with ground truths for patient-level diagnosis, and are comparable with those of experienced doctors.
\section{Discussion}
In this section, we will argue the utility of DeepLung by visualizing the nodule detection and classification results.
\subsection{Nodule Detection}
\begin{figure*}[h]
	\begin{center}
		\includegraphics[width=\linewidth]{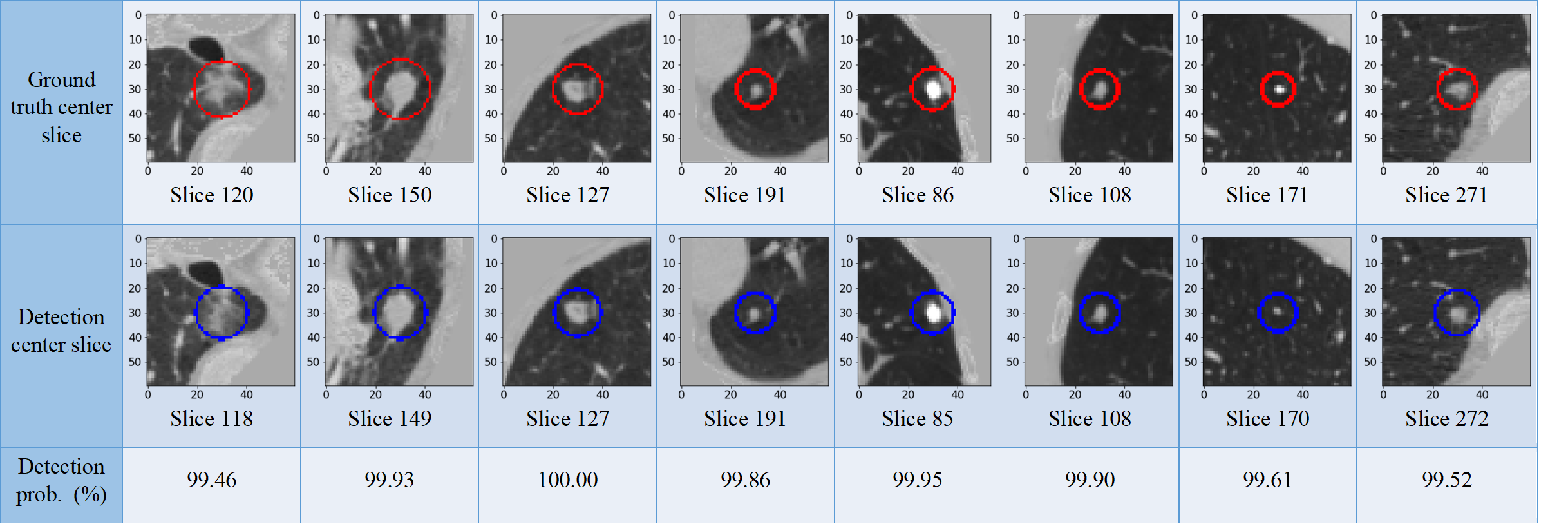}
		\caption{Visualization of central slices for nodule ground truths and detection results. We randomly choose nodules (red circle boxes in the first row) from test fold 1. Detection results are shown in the blue circles of second row. The center slice numbers are shown below the images. The last row shows detection probability. The DeepLung performs well for nodule detection.}
		\label{fig:detection}
	\end{center}
\end{figure*}

We randomly pick nodules from test fold 1 and visualize them in red circles in the first row of Fig. \ref{fig:detection}. Detected nodules are visualized in blue circles of the second row. Because CT is 3D voxel data, we can only plot the central slice for visualization. The third row shows the detection probabilities for the detected nodules. The central slice number is shown below each slice. The diameter of the circle is relative to the nodule size.

From the central slice visualizations in Fig. \ref{fig:detection}, we observe the detected nodule positions including central slice numbers are consistent with those of ground truth nodules. The circle sizes are similar between the nodules in the first row and the second row. The detection probability is also very high for these nodules in the third row. It shows 3D Faster R-CNN works well to detect the nodules from test fold 1. 
\subsection{Nodule Classification}
\begin{figure}[!h]
	\begin{center}
		\includegraphics[width=0.9\linewidth]{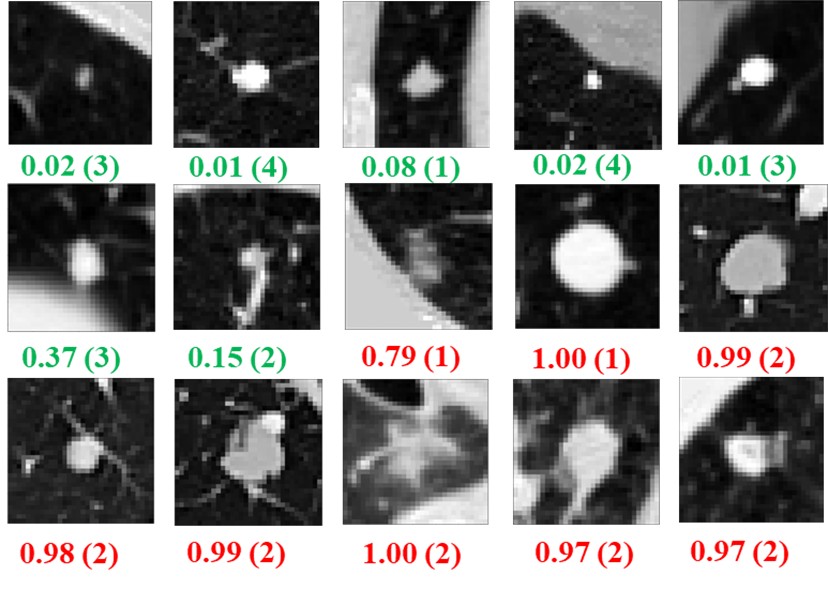}
		\caption{Visualization of central slices for nodule classification results on test fold 1. We choose nodules that are predicted right by the DeepLung, but annotated incorrectly by some doctors. The numbers below the nodules are model predicted malignant probabilities followed by which annotation of doctors is wrong. The first seven nodules are benign nodules. The rest nodules are malignant nodules. The DeepLung performs well for nodule classification.}\label{fig:classification}
	\end{center}
\end{figure}

We also visualize the nodule classification results from test fold 1 in Fig. \ref{fig:classification}. We choose nodules that is predicted right, but annotated incorrectly by some doctors. The first seven nodules are benign nodules, and the remaining nodules are malignant nodules. The numbers below the figures are the DeepLung predicted malignant probabilities followed by which annotation of doctors is wrong. For the DeepLung, if the probability is larger than 0.5, it predicts malignant. Otherwise, it predicts benign. For an experienced doctor, if a nodule is large and has irregular shape, it has a high probability to be a malignant nodule. 

From Fig. \ref{fig:classification}, we can observe that doctors mis-diagnose some nodules. The reason may be be that humans are not fit to process 3D CT data which are of low signal to noise ratio. Perhaps some doctors cannot find some weak irregular boundaries or erroraneously consider some normal tissues as nodule boundaries leading to false negatives or false positives. In addition, doctors' own internal bias may play a role in how confident he/she predicts these scans while being limited to observing only one slice at a time. Machine learning-based methods can overcome these limitations and are able to learn complicated rules and high dimensional features while utilizing all input slices at once without much problem. From this perspective, DeepLung can potentially be of great use to doctors in their effort to make consistent and accurage diagonsis. 

\section{Conclusion}
In this work, we propose a fully automated lung CT cancer diagnosis system based on deep learning. DeepLung consists of two parts, nodule detection and classification. To fully exploit 3D CT images, we propose two deep 3D convolutional networks based on 3D dual path networks, which is more compact and can yield better performance than residual networks. For nodule detection, we design a 3D Faster R-CNN with 3D dual path blocks and a U-net-like encoder-decoder structure to detect candidate nodules. The detected nodules are subsequently fed to nodule classification network. We use a deep 3D dual path network to extract classification features. Finally, gradient boosting machine with combined features are trained to classify candidate nodules into benign or malignant. Extensive experimental results on public available large-scale datasets, LUNA16 and LIDC-IDRI datasets, demonstrate the superior performance of the DeepLung system.

{\small
	\bibliographystyle{ieee}
	\bibliography{egbib}
}

\end{document}